%% file: main.tex
\documentclass[10pt,twocolumn,letterpaper]{article}

\usepackage{iccv}



\usepackage{algorithm}
\usepackage{algorithmic}

\definecolor{iccvblue}{rgb}{0.21,0.49,0.74}
\usepackage[pagebackref,breaklinks,colorlinks,allcolors=iccvblue]{hyperref}

\title{Robust Visual Localization via Semantic-Guided Multi-Scale Transformer}

\author{
Zhongtao Tian\textsuperscript{1,2}$^\ast$ \and 
Wenhao Huang\textsuperscript{1}$^\ast$ \and 
Zhidong Chen\textsuperscript{2}$^\dag$ \and 
Xiao Wei Sun\textsuperscript{1}$^\dag$
\\
\textsuperscript{1}Southern University of Science and Technology \\
\textsuperscript{2}Peng Cheng Laboratory \\
{\tt\small \{tianzt,huangwh\}@mail.sustech.edu.cn, chenzd@pcl.ac.cn, sunxw@sustech.edu.cn}
\\
{\small $^\ast$Equal contribution \quad $^\dag$Corresponding author}
}

\begin{document}
\maketitle
\input{sec/0_abstract}  
\input{sec/1_intro}

\input{sec/2_related}

\input{sec/3_method}

\input{sec/4_experiments}
\input{sec/5_conclusion}

{
    \small
    \bibliographystyle{ieeenat_fullname}
    \bibliography{references}
}

\end{document}

%% file: sec/0_abstract.tex
\begin{abstract}
Visual localization remains challenging in dynamic environments where fluctuating lighting, adverse weather, and moving objects disrupt appearance cues. Despite advances in feature representation, current absolute pose regression methods struggle to maintain consistency under varying conditions.
To address this challenge, we propose a framework that synergistically combines multi-scale feature learning with semantic scene understanding. Our approach employs a hierarchical Transformer with cross-scale attention to fuse geometric details and contextual cues, preserving spatial precision while adapting to environmental changes. We improve the performance of this architecture with semantic supervision via neural scene representation during training, guiding the network to learn view-invariant features that encode persistent structural information while suppressing complex environmental interference.
Experiments on TartanAir demonstrate that our approach outperforms existing pose regression methods in challenging scenarios with dynamic objects, illumination changes, and occlusions. Our findings show that integrating multi-scale processing with semantic guidance offers a promising strategy for robust visual localization in real-world dynamic environments.
\end{abstract}

%% file: sec/1_intro.tex
\section{Introduction}
\label{sec:intro}

Visual localization is essential for numerous applications from robot navigation~\cite{6224766} to augmented reality~\cite{6777443}. The task involves estimating camera position and orientation within a reference coordinate system using only visual information. However, despite significant progress, accurately determining camera pose remains challenging under varying conditions, such as changes in illumination, weather, or the presence of dynamic objects.

Absolute Pose Regression (APR) has emerged as a promising approach for visual localization. These methods directly estimate camera poses from images, offering an alternative to traditional structure-based techniques like Structure-from-Motion (SfM)~\cite{7780814}. The computational efficiency and ability to handle ambiguous features make APR particularly attractive for real-world applications. However, most APR methods perform inconsistently across different environmental conditions.

Environmental variations pose significant challenges for visual localization systems. Recent APR approaches~\cite{8578375,8100177,7410693} have improved feature representation, but maintaining consistent performance across diverse conditions remains difficult~\cite{8578995}. Illumination changes, weather variations, and dynamic objects introduce complex appearance variations that significantly affect pose estimation accuracy. Models trained on daytime images often struggle with nighttime localization due to dramatic changes in lighting conditions and visual features, as demonstrated by recent studies~\cite{9577733,8885732}.
A key limitation is their reliance on appearance-based features processed at a single semantic level. These approaches fail to capture how environmental changes affect features differently across scales. Lighting primarily impacts fine textures while preserving global structure, whereas moving objects alter specific spatial regions. We realized that robust localization requires understanding scene structure across multiple scales and semantic levels, allowing the model to adaptively extract features that remain stable under different conditions.

\begin{figure*}[ht]
  \centering
   \includegraphics[width=0.8\textwidth]{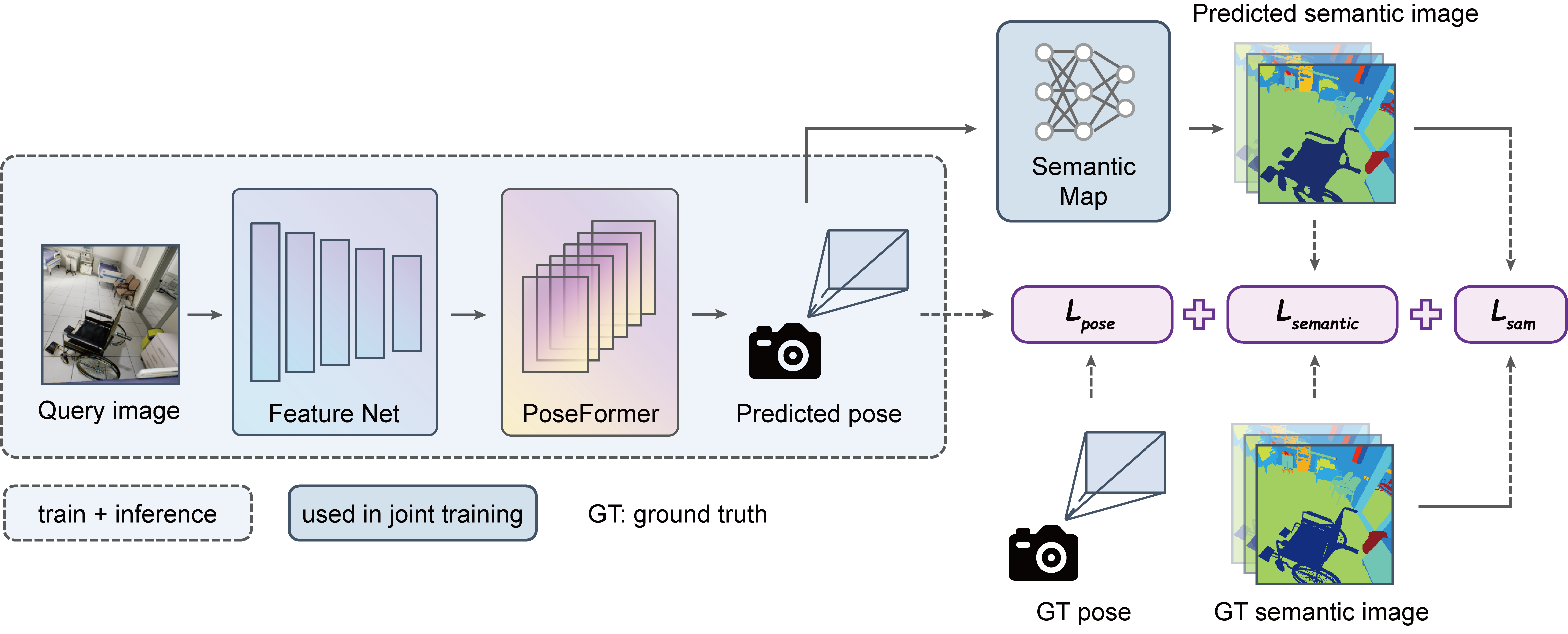}
   \caption{Overview of our framework. The network consists of a feature extraction backbone followed by PoseFormer for pose estimation (left). During joint training, we leverage a semantic map module to generate view-consistent semantic predictions, providing additional supervision through semantic consistency and spectral angle mapping losses (right). Note that the semantic supervision components (shaded in blue) are only utilized during joint training.}
   \label{fig:framework}
\end{figure*}

Neural Radiance Fields (NeRF)~\cite{NeRF} offer new possibilities for visual localization through their ability to represent scenes implicitly~\cite{iNeRF,10.1609/aaai.v38i7.28576,10161117,sparsepose,nerfels}. Recent advances, particularly in approaches like DFNet~\cite{dfnet} and LENS~\cite{pmlr-v164-moreau22a}, have demonstrated significant potential in enhancing APR performance through view synthesis. While NeRF has shown promise in visual localization, its semantic capabilities have yet to be fully leveraged. 
Studies on semantic NeRF~\cite{9710936} have revealed that semantic features remain highly consistent across different viewpoints and are naturally resistant to environmental variations. This property points to promising directions for enhancing localization robustness.

Based on these observations, we propose a framework that aims to address the fundamental challenges of visual localization under changing conditions. Our approach uses an hierarchical Transformer that integrates features across multiple scales, handling diverse environmental challenges from lighting changes to moving objects. 
Unlike existing APR methods that either process features at a single level or rely solely on appearance consistency, we combines multi-scale feature learning with semantic supervision. During development, we found this combination particularly effective: the semantic guidance helps maintain consistency across environmental changes while the multi-scale processing adapts to different types of variations. 

Experiments on the challenging TartanAir~\cite{tartanair} dataset demonstrate the effectiveness of our approach across diverse conditions including scenarios with dynamic objects, lighting changes, and occlusions. Our contributions include:

\medskip

\begin{itemize}[leftmargin=2em]
    \item A multi-scale Transformer architecture that effectively integrates hierarchical features for robust pose estimation.
    
    \medskip
    
\end{itemize}

\begin{itemize}[leftmargin=2em]
    \item The first integration of semantic NeRF supervision in visual localization, enhancing robustness against environmental variations.
    
    \medskip
    
\end{itemize}

\begin{itemize}[leftmargin=2em]
    \item State-of-the-art (SOTA) performance on challenging benchmarks with analysis under various environmental conditions.
    
\end{itemize}

%% file: sec/2_related.tex
\section{Related Work}
\label{sec:related}

\subsection{Absolute Pose Regression (APR)}

Absolute Pose Regression (APR) has emerged as an influential approach in visual localization, offering an elegant solution through end-to-end pose estimation. The seminal work of PoseNet~\cite{7410693} pioneered this direction by adapting a CNN backbone with an MLP regressor for direct pose prediction, demonstrating the feasibility of learning-based camera localization. Following works explored various technical innovations: LSTM layers were introduced for temporal modeling~\cite{8237337}, hourglass networks were adopted for hierarchical feature extraction~\cite{8265316}, and attention mechanisms were incorporated for global context modeling~\cite{10.1007/978-3-031-20080-9_9,atloc}. Kendall et al.~\cite{8100177} proposed learnable weights to balance translation and rotation components, while Bayesian PoseNet~\cite{7487679} incorporated uncertainty estimation through Monte Carlo dropout. MapNet~\cite{mapnet} combined both absolute pose loss and relative pose constraints while maintaining single-frame inference capability, and RelocNet~\cite{relocnet} introduced camera frustum overlap for enhanced feature learning. Recent works have further advanced the field: IRPNet~\cite{irpnet} adopted pre-trained image retrieval models as feature extractors, E-PoseNet~\cite{eposenet} incorporated group-equivalent CNNs for geometric-aware features, and CamNet~\cite{camnet} proposed bilateral frustum loss. While these developments have collectively advanced APR’s capabilities, ensuring consistent localization performance across diverse environmental conditions remains a challenging open research question.
\subsection{NeRF for APR}
The integration of NeRF into visual localization has opened new avenues for enhancing pose estimation accuracy~\cite{direct-pose}. Early applications primarily focused on data augmentation and view synthesis. LENS~\cite{pmlr-v164-moreau22a} utilized pretrained NeRF models to generate synthetic training data. DFNet~\cite{dfnet} introduced online rendering for pose refinement through feature comparison between real and synthesized views.

Recent developments have expanded beyond simple view synthesis to leverage NeRF's implicit scene understanding capabilities. NeRF-loc~\cite{nerf-loc} demonstrated the potential of incorporating implicit 3D descriptors, while CROSSFIRE~\cite{crossfire} introduced novel approaches for 2D-3D feature matching. These advancements highlight NeRF's unique ability to aggregate multi-view information and maintain pose-aware scene representations, suggesting promising directions for further improving localization robustness and accuracy.

%% file: sec/3_method.tex
\section{Method}
\label{sec:method}

Given a set of images and their associated camera poses $\{(I, p)\}$, our framework aims to learn a robust mapping from a query image $I$ to its corresponding camera pose $p$. The camera pose is represented by a translation vector $x$ and a quaternion $q$, which together encode both the translation and rotation relative to a reference coordinate system. 
As illustrated in ~\cref{fig:framework}, our framework consists of two key components: PoseFormer and SemanticMap. PoseFormer employs an hierarchical architecture with cross-scale attention to extract and integrate multi-scale features for pose prediction. Next, SemanticMap leverages the ground-truth pose $p$ to synthesize a semantic image $\hat{I}$, thereby enabling semantic consistency supervision. Notably, the semantic supervision is only required during joint training, allowing efficient inference while maintaining the end-to-end simplicity characteristic of pose regression methods. The rest of this section details our technical contributions: ~\cref{sec:poseformer} presents the hierarchical Transformer-based network architecture for pose regression, ~\cref{sec:semanticmap} introduces the semantic synthesis network SemanticMap, and ~\cref{sec:loss} shows the loss function optimized during joint training.

\subsection{PoseFormer: Multi-scale Feature Learning}
\label{sec:poseformer}
\begin{figure}[ht]
  \centering
   \includegraphics[width=0.7\linewidth]{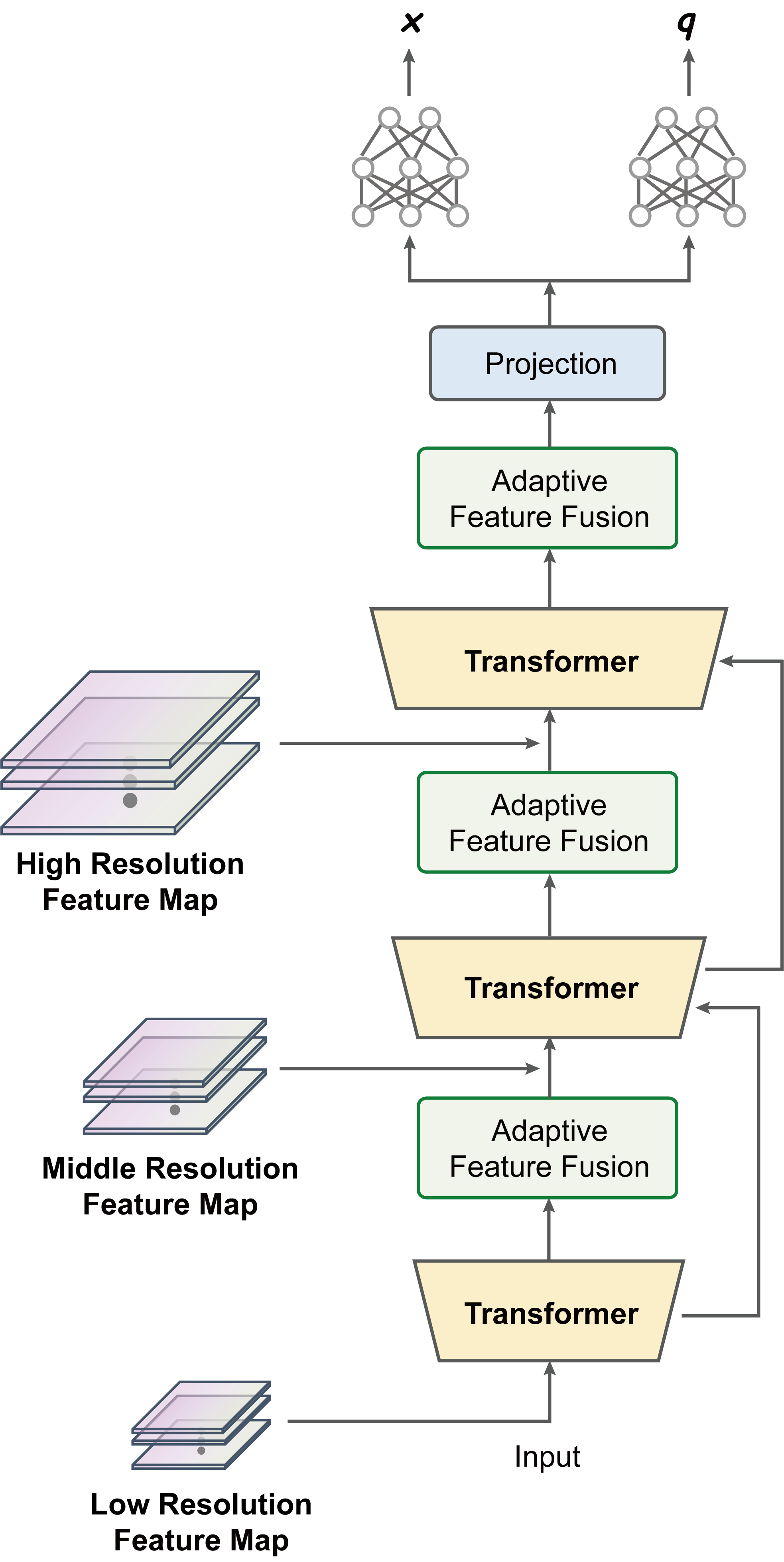}
   \caption{Architecture of our PoseFormer. Multi-scale features are progressively processed through a series of Transformer blocks with cross-scale attention and adaptive feature fusion.}
   \label{fig:poseformer}
\end{figure}

Our PoseFormer adopts a hierarchical architecture that progressively processes and fuses multi-scale features for robust pose estimation. Unlike conventional feature pyramids that lose important spatial details through downsampling operations, our hierarchical design preserves spatial information while integrating features across scales. The network addresses environmental variations by leveraging complementary information in a manner that emphasizes global structure when local features become unreliable due to lighting changes, and prioritizes local details when global cues are diminished. As shown in ~\cref{fig:poseformer}, PoseFormer consists of three main components: multi-scale feature extraction, up-sampling Transformer block, and position-aware feature fusion.

\subsubsection{Multi-scale Feature Extraction}
The backbone network (MobileNetV2~\cite{mobilenetv2}) extracts features at three different scales $\{F_l\}_{l=1}^3$. Specifically, the network produces feature maps at $1/8$, $1/16$, and $1/32$ resolutions with 32, 96, and 1280 channels respectively, enabling comprehensive scene understanding at multiple granularities. To facilitate effective feature fusion, we project these features to a common 256-dimensional space through $1 \times 1$ convolutions. The feature flow follows a bottom-up path where features from the shallowest layer ($1/32$ resolution) are progressively upsampled and fused with higher-resolution features through cross-scale attention. This creates a feature hierarchy where each level integrates information from all previous levels, maintaining both spatial details and semantic context.

\begin{figure*}[ht]
  \centering
   \includegraphics[width=0.8\textwidth]{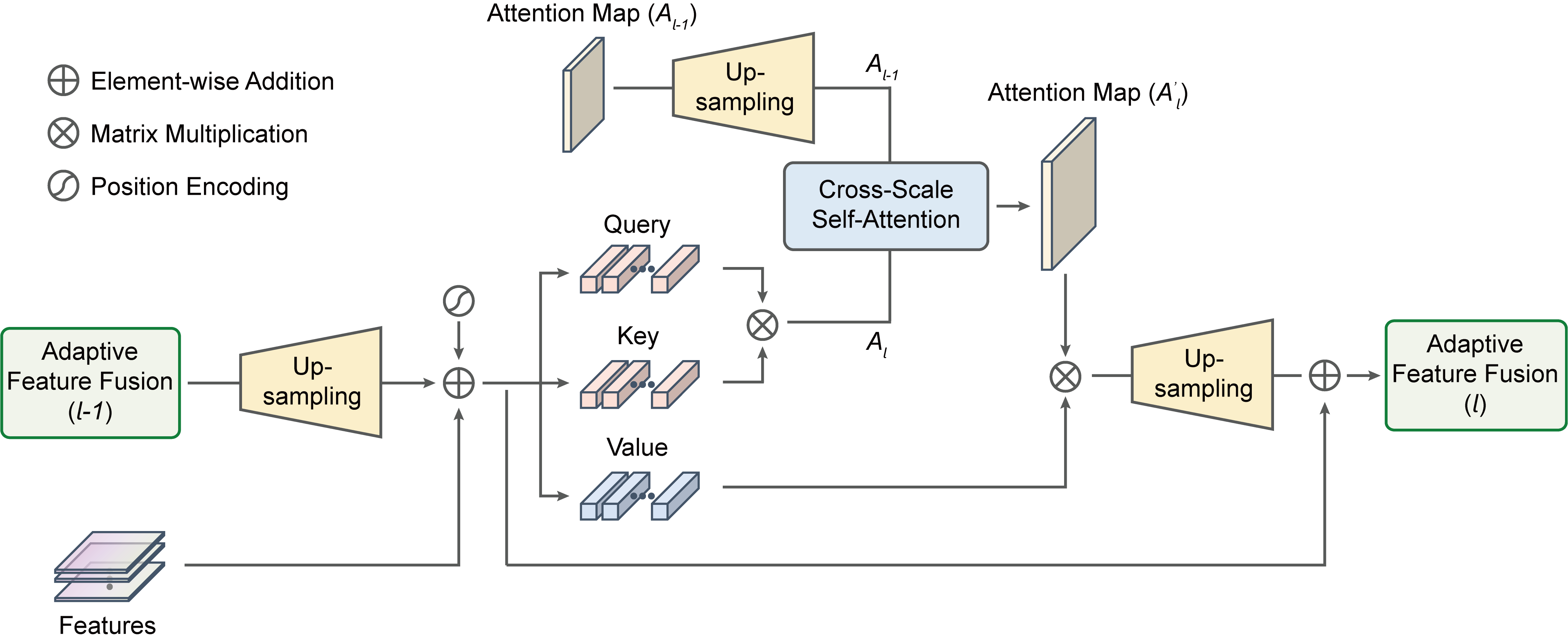}
   \caption{Illustration of the Cross-scale Attention mechanism.}
   \label{fig:cross-scale attention}
\end{figure*}

\subsubsection{Up-sampling Transformer Block}
The up-sampling Transformer serves as the core component of our architecture, designed to progressively integrate multi-scale features while preserving spatial awareness. Given the projected features $F_l^p$, the block first reshapes and upsamples them through a designed two-stage process. 
The features are first reshaped to preserve spatial structure and then upsampled using bilinear interpolation to double the resolution. This is followed by convolution and batch normalization to enhance feature representation. By progressively incorporating semantic information from lower resolutions, this process helps maintain high-resolution spatial details. The resulting upsampled features serve as a rich foundation for subsequent cross-scale attention mechanisms, ensuring the preservation of local details while effectively integrating global context.

\subsubsection{Cross-scale Attention}
Cross-scale feature interaction significantly improves scene representation by capturing both fine-grained details and global context~\cite{crossvit,crossformer,inverted}. This capability is advantageous for visual localization, as accurate pose estimation relies on both precise geometric information and a comprehensive understanding of scene structure.

Our cross-scale attention mechanism addresses a fundamental challenge: different environmental variations affect features at different scales non-uniformly. For instance, illumination changes primarily alter local appearance while preserving global structure, whereas dynamic objects mainly affect specific spatial regions. By enabling content-adaptive interaction between features across scales, our mechanism learns to selectively emphasize the most reliable scale-specific features under different conditions. For each scale level $l$, features are first projected into query, key, and value spaces:
\begin{equation}
    Q_l, K_l, V_l = \text{proj}_q(F_l), \text{proj}_k(F_l), \text{proj}_v(F_l),
\end{equation}
The attention computation uses a progressive fusion strategy:
\begin{align}
    A_l &= \frac{Q_lK_l^T}{\sqrt{d_k}}, \\
    A_l^{'} &= \omega_{l,1}A_l + \omega_{l,2}A_{l-1},
\end{align}
where $A_l$ represents the attention map at current scale, $A_{l-1}$ denotes the attention from previous scale, and $\omega_{l,1}, \omega_{l,2}$ are learnable fusion weights that adaptively balance their contributions. This adaptive weighting strategy enhances the network's robustness to diverse environmental variations without requiring explicit modeling of each variation type.

\subsubsection{Position-aware Feature Fusion}
To maintain precise spatial awareness crucial for pose estimation, we incorporate a multi-level position encoding scheme that adaptively combines positional information with feature representations. The position encoding follows a sinusoidal pattern with learnable scale factors, allowing the network to adjust the importance of positional information at different scales. 
\begin{equation}
    \text{PE}_l(p,c) = 
    \begin{cases}
        \sin\left(\frac{p}{10000^{2c/d}}\right), & \text{if } c \text{ is even} \\
        \cos\left(\frac{p}{10000^{2c/d}}\right), & \text{if } c \text{ is odd}
    \end{cases}
\end{equation}

\begin{equation}
    F_l^{\text{pe}} = F_l^{\text{up}} + \alpha_l \cdot \text{PE}_l(H_l, W_l),
\end{equation}
where $\alpha_l$ is a learnable scale factor that allows the network to adaptively adjust the importance of positional information. 
During training, the network learns to regulate $\alpha_l$ based on the reliability of positional cues under different conditions, emphasizing positional information in environments with distinctive spatial layouts while relying more on semantic features in repetitive or ambiguous environments.

\subsubsection{Dual-head Pose Regression}
The final pose estimation employs a dual-head architecture that separately processes translation and rotation while maintaining their correlation through shared features. Separating these tasks allows each head to focus on task-specific features while mitigating the inherent scale difference between translation and rotation errors that can destabilize training. Each head employs a progressive refinement network that gradually transforms the 1280-dimensional features through multiple stages of dimensionality reduction. The translation head outputs a 3-dimensional position vector, while the rotation head produces a 4-dimensional quaternion representation.

\subsection{SemanticMap: Neural Scene Supervision}
\label{sec:semanticmap}

The key insight behind our semantic neural rendering lies in its ability to provide view-invariant supervision signals that remain stable under challenging environmental conditions. Unlike appearance-based supervision that is highly sensitive to illumination changes and dynamic objects, semantic labels maintain consistency across different conditions. This semantic consistency coupled with the inherent 3D structure awareness from volumetric rendering, enables more reliable pose estimation in complex environments.

We extend NeRF to incorporate semantic supervision for pose estimation. Unlike standard NeRF that models appearance, we represent semantic segmentation as a view-invariant function mapping 3D coordinates to semantic classes. For each 3D point $\mathbf{x}$ and viewing direction $\mathbf{d}$, our network outputs:
\begin{equation}
    \mathbf{s} = F_\Theta(\mathbf{x})
\end{equation}
where $F_\Theta$ represents the learned MLP and $\mathbf{s} \in \mathbb{R}^C$ indicates semantic logits over $C$ categories. Notably, the semantic prediction depends only on spatial coordinates, reflecting the view-invariant nature of semantics.

The semantic rendering follows volumetric integration along camera rays. For a ray $r(t) = \mathbf{o} + t\mathbf{d}$, we sample $N$ points and compute the expected semantic logits:
\begin{equation}
    S(r) = \sum_{i=1}^N T_i \left(1 - \exp(-\sigma_i \delta_i)\right) \mathbf{s}_i,
\end{equation}
where $T_i = \exp\left(-\sum_{j<i} \sigma_j \delta_j\right)$ represents accumulated transmittance, and $\delta_i$ denotes the distance between adjacent samples. Importantly, we only need this semantic supervision during joint training. At inference time, our model directly estimates poses from RGB images, maintaining computational efficiency comparable to standard APR methods.

\subsection{Loss Function}
\label{sec:loss}

Our training objective combines pose regression and semantic supervision:
\begin{equation}
    \mathcal{L} = \lambda_p \mathcal{L}_{\text{pose}} + \lambda_s \mathcal{L}_{\text{semantic}}
\end{equation}

\subsubsection{Pose Regression Loss}
For pose regression, we train our model to minimize both position and orientation errors with respect to ground truth pose $p_0 = \{x_0, q_0\}$. The position loss $\mathcal{L}_x$ and orientation loss $\mathcal{L}_q$ are given by:
\begin{align}
\mathcal{L}_x &= \|x_0 - x\|_2 \\
\mathcal{L}_q &= \|q_0 - \frac{q}{\|q\|}\|_2
\end{align}
where the predicted quaternion $q$ is normalized to ensure valid orientation encoding. Following Kendall et al.~\cite{8100177}, we combine these losses using learned parameters to automatically balance their contributions:
\begin{equation}
\mathcal{L}_{\text{pose}} = \mathcal{L}_x \exp(-s_x) + s_x + \mathcal{L}_q \exp(-s_q) + s_q
    \label{eq:poss_loss}
\end{equation}
where $s_x$ and $s_q$ are learnable parameters that control the relative weighting between position and orientation components during training.

\subsubsection{Semantic Supervision Loss}
The semantic supervision employs a dual-component loss function:
\begin{equation}
    \mathcal{L}_{\text{semantic}} = \mathcal{L}_{\text{ce}}(S, S_{\text{gt}}) + \lambda_{\text{sam}} \mathcal{L}_{\text{sam}}(S, S_{\text{gt}}).
    \label{eq:semantic_loss}
\end{equation}
The cross-entropy loss $\mathcal{L}_{\text{ce}}$ ensures semantic class consistency, while the spectral angle mapping loss $\mathcal{L}_{\text{sam}}$ evaluates structural similarity between predicted and ground-truth semantic maps. This combination effectively captures both categorical and structural aspects of scene understanding, enabling the network to learn robust semantic features that generalize well across viewpoints.

%% file: sec/4_experiments.tex
\section{Experiments}
\label{sec:experiments}

\begin{table*}[ht]
\centering
\small
\begin{tabular}{@{}l|ccccc|cc@{}}
\toprule

\multicolumn{6}{c|}{\textbf{Single-frame APR}} & \multicolumn{2}{c}{\textbf{APR with Unlabelled Data}} \\
\cmidrule(r){1-6} \cmidrule(r){7-8}
\textbf{Scenes} & PoseNet & Direct-PN & MS-T & DFNet & \textbf{PoseFormer} & Direct-PN+U & \textbf{PoseFormer$_{sem}$} \\

\midrule
Hospital & 0.47/6.72 & 0.26/3.31 & 0.28/4.11 & 0.27/2.61 & \textbf{0.14/1.37} & 0.21/2.32 & \textbf{0.14/1.22} \\
Factory\_Day & 0.93/9.62 & 0.35/4.36 & 0.33/4.21 & 0.25/2.45 & \textbf{0.19/0.96} & 0.28/2.71 & \textbf{0.19/0.96} \\
Factory\_Night & 0.75/10.35 & 0.35/3.63 & 0.35/4.79 & 0.30/2.24 & \textbf{0.21/0.87} & 0.35/5.02 & \textbf{0.20/0.81}  \\
Office & 0.61/5.81 & 0.39/3.26 & 0.39/2.84 & 0.21/2.19 & \textbf{0.16/0.84} & 0.29/2/87 & \textbf{0.16/0.78} \\
Winter & 0.83/10.64 & 0.44/3.10 & 0.42/3.91 & 0.40/2.19 & \textbf{0.33/0.80} & 0.42/3.93 & \textbf{0.30/0.81} \\
Japanesealley & 0.58/5.99 & 0.31/4.16 & 0.33/2.57 & 0.21/1.26 & \textbf{0.17/0.52} & 0.32/4.54 & \textbf{0.15/0.49} \\
Seasonsforest & 0.58/5.94 & 0.30/3.00 & 0.30/3.70 & 0.25/1.69 & \textbf{0.18/0.94} & 0.28/2.82 & \textbf{0.17/0.81} \\
Carwelding & 0.70/9.24 & 0.43/4.15 & 0.44/3.32 & \textbf{0.28}/1.44 & \textbf{0.28/0.76} & 0.41/4.36 & \textbf{0.27/0.72} \\
Amusement & 0.76/12.23 & 0.56/2.63 & 0.30/2.95 & 0.31/1.72 & \textbf{0.29/0.82} & 0.34/1.58 & \textbf{0.26/0.81} \\
Neighborhood & 1.07/5.29 & 0.38/4.70 & 0.59/3.39 & 0.30/3.05 & \textbf{0.26/0.85} & 0.36/3.24 & \textbf{0.26/0.83} \\
Soulcity & 0.62/11.82 & 0.47/3.12 & 0.44/3.46 & 0.35/2.00 & \textbf{0.29/0.61} & 0.39/3.32 & \textbf{0.28/0.57} \\
Westerndesert & 0.59/11.16 & 0.41/2.86 & 0.39/3.71 & 0.37/2.14 & \textbf{0.33/0.87} & 0.40/3.02 & \textbf{0.31/0.78} \\
Seasidetown & 0.71/10.34 & 0.36/4.54 & 0.34/3.63 & 0.26/2.53 & \textbf{0.17/1.18} & 0.34/3.80 & \textbf{0.17/1.01} \\
Gascola & 0.47/7.78 & 0.28/4.15 & 0.29/3.90 & \textbf{0.19}/1.42 & \textbf{0.19/0.86} & 0.26/3.74 & \textbf{0.18/0.82} \\

\midrule
Average & 0.69/8.78 & 0.38/3.64 & 0.37/3.61 & 0.28/2.07 & \textbf{0.23/0.88} & 0.33/3.38 & \textbf{0.22/0.82} \\

\bottomrule
\end{tabular}
\caption{Pose regression results on TartanAir dataset. We report the average of median translation and rotation error in m/$^\circ$. The best results are highlighted in \textbf{bold}.}
\label{tab:results}
\end{table*}

\subsection{Datasets and Details}
\textbf{Datasets.} We evaluate our approach on the TartanAir~\cite{tartanair} dataset, a large-scale synthetic benchmark specifically designed to challenge visual localization algorithms in complex environments. This dataset is particularly suitable for evaluating our method's robustness to environmental variations due to its diverse scenes with changing lighting conditions, adverse weather, and complex geometries. For our comparative analysis, we consider fourteen scenes that cover urban, rural, natural, indoor, and outdoor environments. Some of these scenes have time-varying lighting (day/night transitions), overexposure and dark lighting, adverse weather (rain), and low-texture conditions (snowstorms). Each scene spans a wide spatial range with aggressive 6-DOF camera motions, simulating trajectories up to 100 m $\times$ 100 m $\times$ 10 m in challenging layouts. For all studies, we use the ground-truth image-pose pairs for training.

\textbf{Training Details.}
The overall training process is divided into three stages, each serving a specific purpose in our learning framework.

\textbf{Stage 1: Pose Regression Pretraining.}
We initialize PoseFormer with an ImageNet-pretrained MobileNetV2~\cite{mobilenetv2} backbone and train it using pose supervision in~\cref{eq:poss_loss} for 2000 epochs. This stage establishes a strong foundation for geometric understanding through direct pose regression. The loss is optimized via Adam ($\beta_1=0.9$, $\beta_2=0.999$, $\epsilon=10^{-10}$) with an initial learning rate of $10^{-4}$, weight decay of $10^{-4}$, and batch size of 8. Input images are resized to $320 \times 240$ to balance computational efficiency and feature resolution. Learning rate scheduling is governed by ReduceLROnPlateau, reducing the rate by 5\% after 50 epochs of validation stagnation, while early stopping (patience=200 epochs) terminates training if no improvement in the validation set.

\textbf{Stage 2: Semantic Neural Rendering.}
We adopt the Semantic-NeRF~\cite{9710936} architecture for neural scene representation learning, but simplify its model by using only a coarse MLP (omitting the fine MLP used in vanilla NeRF and DFNet) to improve training efficiency. The model is optimized via Adam (initial lr=$0.001$) with an exponential decay schedule. Training runs for 4,000 epochs with a loss combining RGB loss and semantic cross-entropy loss (weight=$4 \times 10^{-2}$). This stage creates a semantic neural scene representation that will provide view-invariant supervision signals.

\textbf{Stage 3: Semantic-guided Refinement.}
We freeze the backbone network of PoseFormer while unfreezing batch normalization layers to adapt to NeRF-generated features. We disable direct pose supervision and instead rely solely on the semantic consistency loss defined in~\cref{eq:semantic_loss} with weights $[0.7, 0.3]$ for cross-entropy and spectral angle mapping components, respectively. This design choice encourages the network to learn view-invariant features that disentangle static scene geometry from dynamic environmental changes. 
Training uses Adam with a lower initial learning rate ($10^{-5}$) and the loss from~\cref{eq:semantic_loss}. All experiments are conducted on a 24GB NVIDIA GeForce RTX 3090 GPU.

\subsection{Evaluation}
\label{sec:evaluation}
We compared the performance of our method with previous single-frame APR methods and unlabeled training approaches in~\cref{tab:results}. The results show that both our PoseFormer and PoseFormer$_{sem}$ (with semantic supervision) achieve superior accuracy across diverse environments. Specifically, PoseFormer achieves 18\% improvement in translation error (0.23 m vs. 0.28 m) and 57\% improvement in median rotation error (0.88$^\circ$ vs. 2.07$^\circ$) compared to the previous SOTA method (DFNet).
Our evaluation encompasses a diverse array of challenging scenes, including both indoor and outdoor environments. The performance improvements are significant in scenes with complex environmental variations. For instance, in the overexposed `Factory\_Day' scene, our method reduces rotation error by 61\% compared to DFNet. Similarly, in the low-light `Factory\_Night' scene, we achieve a 30\% reduction in translation error and a 61\% reduction in rotation error.
Notably, in challenging conditions with low texture, such as `Winter' and `Neighborhood', all methods experience some performance degradation. However, our method maintains robust performance in these complex scenarios. This demonstrates the effectiveness of our multi-scale feature fusion approach in leveraging complementary information across different scales when certain feature types become unreliable.
The semantic-supervised variant (PoseFormer$_{sem}$) shows additional improvements, further validating the value of semantic guidance in enhancing feature robustness.

\begin{figure}[t]
  \centering
   \includegraphics[width=0.95\linewidth]{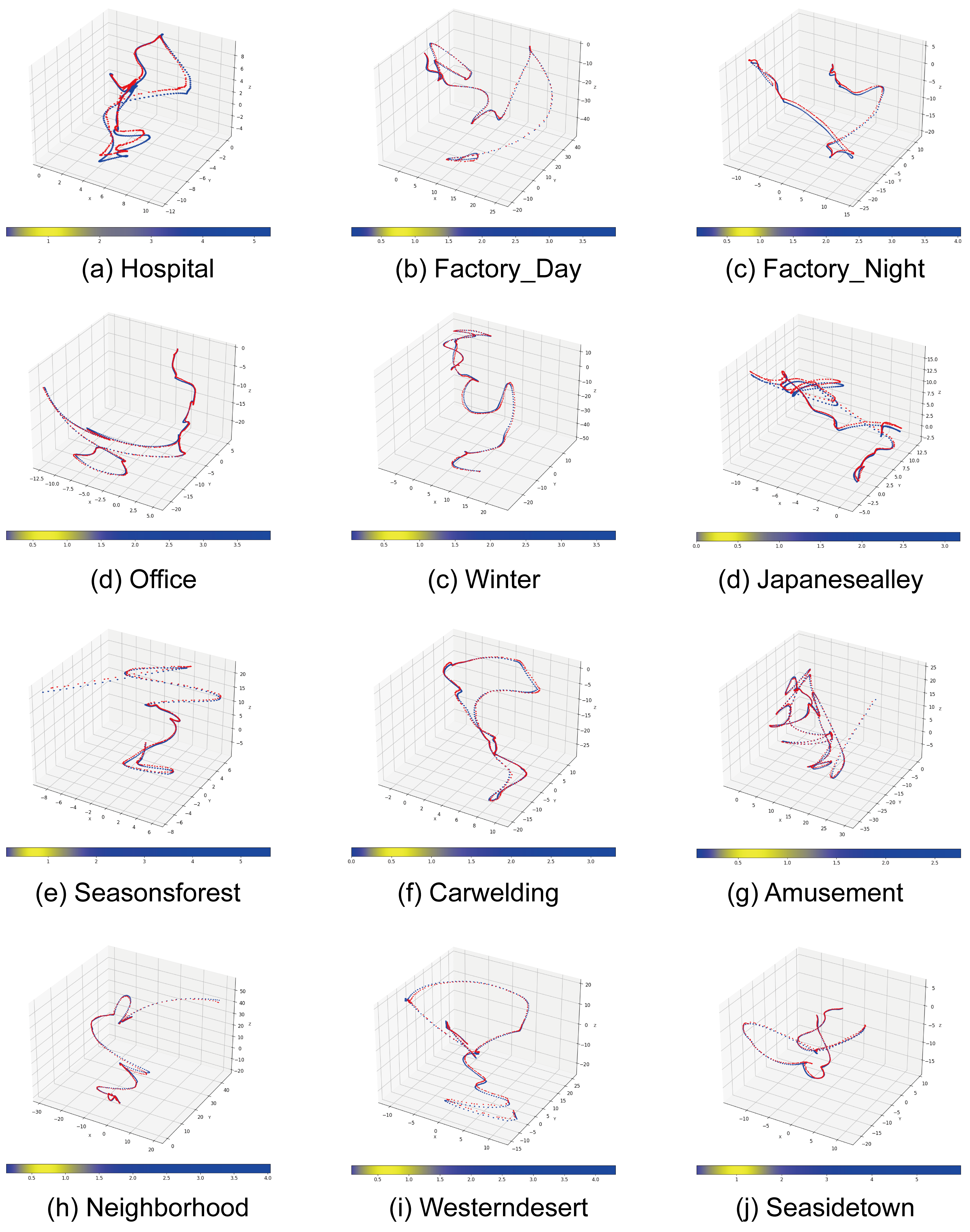}
   \caption{Visual comparison of camera localization between the ground truth (GT) pose and our method on the TartanAir dataset. Each sub-figure presents the GT camera trajectory (blue) with the estimated trajectory (red). The color bar below each plot represents the distribution of rotation errors, where warmer colors (yellow) indicate regions with higher error frequency, while cooler colors (blue) denote areas with lower error frequency.}
   \label{fig:results1}
\end{figure}

We visualize the camera localization sequences on the TartanAir dataset in~\cref{fig:results1}, demonstrating that our estimated trajectories closely follow the ground truth, suggesting high localization accuracy. The majority of rotation errors fall within a small range (1.5$^\circ$), with only a limited number of high-error regions.

\subsection{Attention Maps Visualization and Analysis}
\begin{figure}[t]
  \centering
   \includegraphics[width=0.95\linewidth]{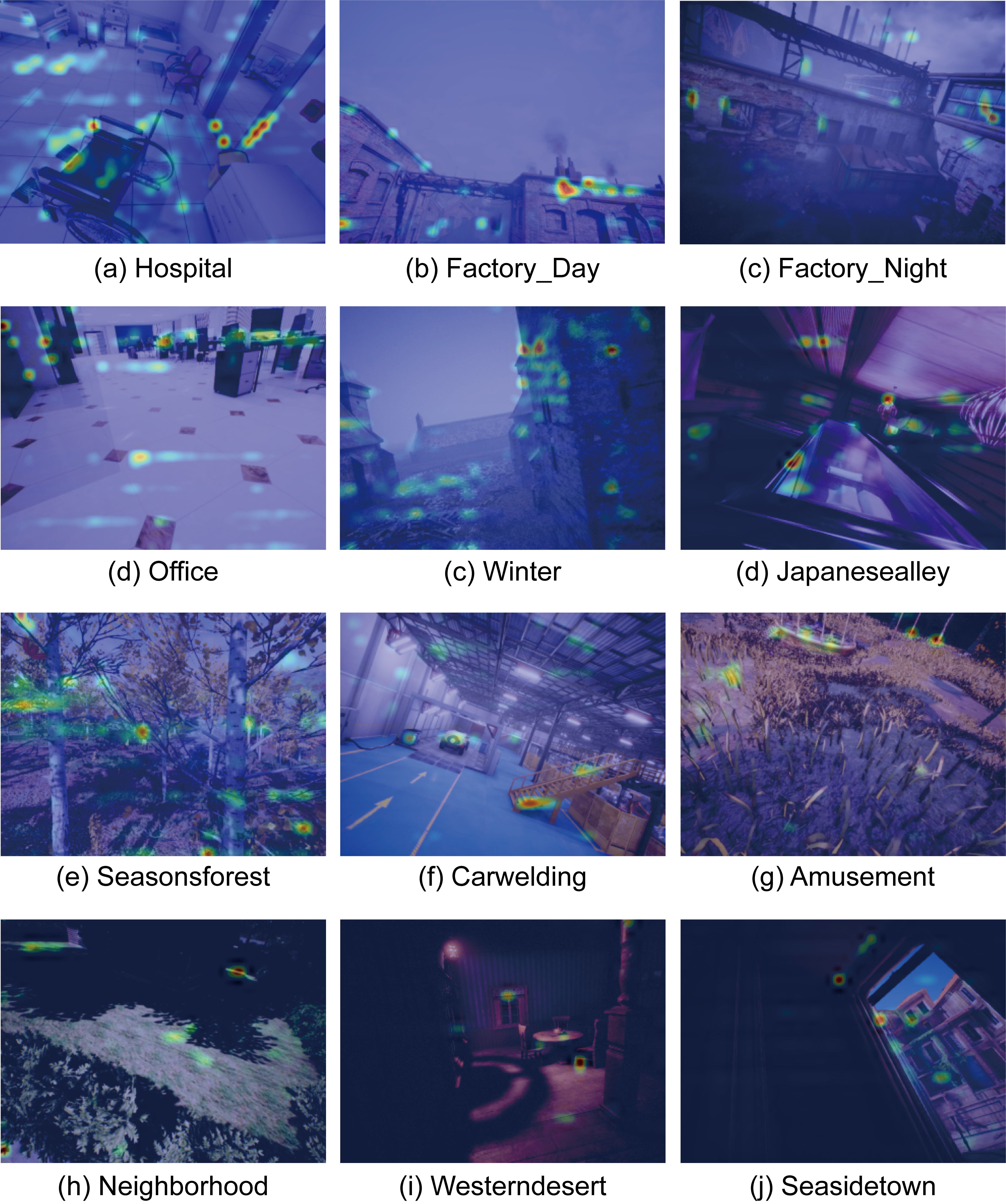}
   \caption{Overlaid attention maps for 12 different scenes, encompassing both indoor and outdoor environments. The visualizations show that our model consistently attends to key localization cues across varying conditions.}
   \label{fig:results2}
\end{figure}

In our experiments, visualizing attention maps serves as a critical tool for interpreting the inner workings of the model and understanding which regions of the input contribute most significantly to the final prediction. As shown in~\cref{fig:results2}, overlaying the attention maps onto the original inputs generates a composite image that highlights the spatial areas where the model's focus is most pronounced.

The attention visualization reveals three characteristic selection patterns: concentrated peaks around distinctive features, moderate activation across textured surfaces, and broad diffusion in homogeneous regions. These spatial distributions directly reflect the model's decision rationale, with clustered high activations typically aligning with critical structural elements essential for recognition. When environmental changes degrade local feature reliability (such as low-light conditions in scene c, i), the attention mechanism automatically shifts focus to more stable structural patterns. Moreover, attention maps exhibit precise geometric localization in structured environments (f, j), while dynamically expanding to capture multi-scale complementary cues in complex scenes (d, e). The correlation between hierarchical attention patterns and pose accuracy empirically validates our architectural design, demonstrating how cross-scale attention fusion enables robust pose estimation across diverse environmental conditions.

\subsection{Ablation Studies} 
To evaluate the impact of different architectural design choices, we conducted a series of ablation studies on the TartanAir dataset (~\cref{tab:ablation_study1} and~\cref{tab:ablation_study2}). Each experiment is based on the architecture used in our comparative analysis in~\cref{sec:evaluation}, with only one algorithmic component modified at a time. We calculate the median position and orientation errors for each scene and report the average metrics. 

\begin{table}[t]
\centering
\begin{tabular}{@{}lccc@{}}
\toprule
\textbf{Backbone} & \textbf{Position} & \textbf{Orientation} \\
\midrule
ResNet18 & 0.32 & 2.24 \\
EfficientNetB0 & 0.27 & 1.39 \\
\textbf{MobileNetV2} & 0.25 & 0.90 \\
\bottomrule
\end{tabular}
\caption{Ablations of the convolutional backbone in our model, evaluated on the TartanAir dataset. We report the average of median position and orientation errors in m/$^\circ$ across all scenes. The chosen model is highlighted in \textbf{bold}.}
\label{tab:ablation_study1}
\end{table}

\begin{table}[t]
\centering
\begin{tabular}{@{}lccc@{}}
\toprule
\textbf{Backbone} & \textbf{Position} & \textbf{Orientation} \\
\midrule
Frozen-23 & 0.31 & 1.32 \\
Frozen-3 & 0.27 & 0.99 \\
\textbf{Full-Train} & 0.25 & 0.90 \\
\bottomrule
\end{tabular}
\caption{Ablations of the multi-scale hierarchical layers of our model, evaluated on the TartanAir dataset. We report the average of median position and orientation errors in m/$^\circ$ across all scenes. The model choice is highlighted in \textbf{bold}.}
\label{tab:ablation_study2}
\end{table}

\textbf{Convolutional Backbone.} We evaluate three convolutional backbones for our pose regression framework: ResNet18, EfficientNetB0, and MobileNetV2, and the results are shown in ~\cref{tab:ablation_study1}. Although ResNet18 (11.7M parameters) is a widely used architecture with proven feature extraction capabilities, its conventional design appears less suited for the spatial precision required in pose regression. EfficientNetB0’s (5.3M parameters) compound scaling mechanism balances accuracy and efficiency but introduces subtle overfitting tendencies under limited training data. 
The lightweight architecture of MobileNetV2 (3.4M parameters) employs inverted residual blocks and linear bottlenecks to prioritize spatially discriminative feature encoding, which is a critical requirement for pose estimation where localized geometric cues outweigh purely semantic understanding. By leveraging these design elements, our model achieves a compact yet expressive feature representation, effectively addressing the challenges of limited training data and optimization complexity.

\textbf{Multi-scale attention architecture.} Our ablation study on multi-scale attention fusion (~\cref{tab:ablation_study2}) reveals progressive performance improvements as deeper hierarchical features are activated. In the Frozen-23 configuration, only the shallowest layer is trainable, resulting in elevated errors due to insufficient handling of global geometric consistency. This limitation is particularly evident in scenes with repetitive structures or textureless regions, where local features alone are insufficient for visual localization. 
Unfreezing the middle layer (Frozen-3) reduces errors by enabling mid-level features to establish spatial dependencies between localized keypoints. This helps correlate transient visual cues (e.g., edge intersections) into more stable regional constraints, improving the model's ability to maintain consistent pose estimates across frames.
The full multi-scale configuration (Full-Train) achieves optimal accuracy, where the deepest layer's high-resolution (30 $\times$ 40) attention maps inject fine-grained geometric details through pixel-wise correlation, while the shallowest layer's low-resolution attention provides global contextual priors by filtering ambiguous spatial patterns. 

This hierarchical design effectively implements a coarse-to-fine pose regression strategy: shallower layers establish global scene understanding and rough pose estimates, while deeper layers progressively refine pose residuals through high-resolution feature matching. Our dynamic fusion mechanism adaptively balances global structural cues for rotation estimation with fine-grained local features for position refinement, naturally decomposing the pose regression into complementary spatial tasks.

%% file: sec/5_conclusion.tex
\section{Conclusion}
\label{sec:conclusion}

In this work, we proposed a novel framework for visual localization in dynamic environments by integrating multi-scale feature learning with semantic scene understanding. Our hierarchical Transformer with cross-scale attention mechanism adaptively fuses features across scales, preserving spatial precision while gaining robustness to environmental variations. The semantic supervision through neural scene representation guides the network to disentangle static geometry from transient changes. Our approach provide a state-of-the-art localization accuracy on the TartanAir benchmark across challenging conditions with dynamic objects, illumination changes, and occlusions, establishing a new direction for robust visual localization in real-world dynamic settings.